\documentclass[conference]{IEEEtran}
\IEEEoverridecommandlockouts

\usepackage{cite}
\usepackage{algorithm} 
\usepackage{algorithmic}  
\usepackage[algo2e]{algorithm2e} 
\usepackage{amsmath,amssymb,amsfonts}
\usepackage{algorithm}
\usepackage{graphicx}
\usepackage{textcomp}
\def\BibTeX{{\rm B\kern-.05em{\sc i\kern-.025em b}\kern-.08em
    T\kern-.1667em\lower.7ex\hbox{E}\kern-.125emX}}
    
    \usepackage{lineno}
\usepackage{lettrine}

\usepackage{url}
\usepackage{moreverb}

\usepackage{graphicx}
\usepackage{subcaption}
\usepackage{float}
\usepackage{amsmath}
\usepackage{array}
\usepackage{multirow}
\graphicspath{ {Figures/} }

\usepackage{listings}
\usepackage{scalerel}
\usepackage{tikz}
\usetikzlibrary{svg.path}

\definecolor{orcidlogocol}{HTML}{A6CE39}
\tikzset{
  orcidlogo/.pic={
    \fill[orcidlogocol] svg{M256,128c0,70.7-57.3,128-128,128C57.3,256,0,198.7,0,128C0,57.3,57.3,0,128,0C198.7,0,256,57.3,256,128z};
    \fill[white] svg{M86.3,186.2H70.9V79.1h15.4v48.4V186.2z}
                 svg{M108.9,79.1h41.6c39.6,0,57,28.3,57,53.6c0,27.5-21.5,53.6-56.8,53.6h-41.8V79.1z M124.3,172.4h24.5c34.9,0,42.9-26.5,42.9-39.7c0-21.5-13.7-39.7-43.7-39.7h-23.7V172.4z}
                 svg{M88.7,56.8c0,5.5-4.5,10.1-10.1,10.1c-5.6,0-10.1-4.6-10.1-10.1c0-5.6,4.5-10.1,10.1-10.1C84.2,46.7,88.7,51.3,88.7,56.8z};
  }
}

\newcommand\orcidicon[1]{\href{https://orcid.org/#1}{\mbox{\scalerel*{
\begin{tikzpicture}[yscale=-1,transform shape]
\pic{orcidlogo};
\end{tikzpicture}
}{|}}}}

\usepackage{hyperref} 

\usepackage{cite}
\usepackage{amsmath,amssymb,amsfonts}
\usepackage{algorithmic}
\usepackage{graphicx}
\usepackage{textcomp}
\usepackage{xcolor}
\usepackage{booktabs}
\usepackage{tabularx}
\usepackage{float} 
\usepackage{hyperref}
\hypersetup{
    colorlinks=true,
    linkcolor=blue,
    citecolor=blue,
    urlcolor=blue
}

\begin{document}

\title{Implementation of Tightly-Coupled SLAM Fusion of GPS, IMU, and LiDAR for Autonomous Vehicles}


\author{Amr O. Elmehrath$^{1,2}$,  Farah Khaled$^{1,2\orcidicon{0009-0001-6869-1288}\,}$, Rana Nahas$^{1,2}$, and Catherine~M.~Elias$^{1,2\orcidicon{0000-0002-1444-9816}\,}$,~\IEEEmembership{Member,~IEEE,}%
\thanks{*This work was not supported by any organization}
\thanks{$^{1}$C-DRiVeS Lab: Cognitive Driving Research in Vehicular Systems, Cairo, Egypt
{\tt\small cdrives.researchlab@gmail.com}}%
\thanks{$^{2}$Computer Science and Engineering Department - Faculty of Media Engineering and Technology - German University in Cairo, Egypt}%
\thanks{{\tt\small farahkhaled03@gmail.com, catherine.elias@ieee.org}}%
}

\maketitle

\begin{abstract}
Autonomous vehicles depend entirely on Simultaneous Localization and Mapping (SLAM) to navigate safely in unknown environments. However, relying on a single sensory modality introduces critical failure points: LiDAR systems degrade in featureless corridors, Inertial Measurement Units (IMUs) accumulate mathematical drift, and GPS drops frequently in urban canyons. This paper presents the implementation of a tightly-coupled SLAM fusion architecture that integrates a Velodyne 3D LiDAR, a high-frequency IMU, and GPS to achieve continuous spatial awareness. Utilizing a phased development methodology, we establish a 2D baseline to validate hardware synchronization and transform geometries before upgrading to a full 3D architecture driven by FAST-LIO2. This advanced approach uses an Iterated Error-State Kalman Filter (IESKF) to process dense 3D laser points alongside continuous inertial data, eliminating motion blur at high speeds. To eradicate long-term drift, a GTSAM pose-graph optimization back-end executes multi-modal loop closures. Evaluated across simulated environments and physical deployments, the results demonstrate that tightly-coupled 3D fusion effectively overcomes individual sensor blind spots to generate highly detailed point clouds, providing the foundational High-Definition (HD) maps required for advanced downstream autonomous planners.
\end{abstract}


\section{Introduction \& Literature Review}
Any autonomous moving machine must constantly answer two fundamental questions: Where am I, and what does the world around me look like? On their own, localization and mapping are well-understood tasks. The challenge arises when a vehicle enters an unknown environment with neither a prior map to localize against nor a known position from which to begin mapping. Solving this circular dependency—where accurate localization needs an accurate map, but building an accurate map needs accurate localization—is the core of Simultaneous Localization and Mapping (SLAM) \cite{yue2024lidar}.

Modern autonomous systems lean heavily on SLAM to provide a continuous, accurate picture of their surroundings. This spatial awareness is not merely for visualization; it is the foundational layer for the entire autonomous vehicle stack. Specifically, SLAM is critical for the planning subsystem. Advanced autonomous planners—such as those utilizing complex agentic behavior trees—strictly require high-definition, SLAM-generated maps to safely and effectively navigate dynamic environments \cite{prompts_pavement}. 

Every individual sensor, however, possesses inherent vulnerabilities. A sensor that maps by measuring distance, such as LiDAR, struggles to extract features in highly symmetrical environments like featureless corridors. An inertial sensor (IMU) tracks motion at high frequencies but accumulates mathematical drift exponentially over time. Global positioning systems (GPS) provide absolute global anchors but lose signal integrity in urban canyons \cite{zhang2025towards}. A vehicle that must remain reliable across open roads, tight urban streets, and featureless interiors cannot trust a single source. 

This paper addresses these vulnerabilities by proposing a tightly-coupled fusion of LiDAR, IMU, and GPS. By feeding raw data into a single shared estimator rather than letting each sensor guess alone, the system covers the blind spots of individual modalities.

Multi-sensor SLAM architectures dictate how measurements from disparate sensors are combined. Early fusion paradigms relied on loose coupling \cite{hening20173d}, where each sensor runs its own independent estimator and hands over a finished trajectory. This structure struggles with real-time error correction, as a failed scan match cannot be rescued by IMU prediction. 

Modern architectures have shifted toward tight coupling \cite{zhou2024tightly}. By feeding raw outputs directly into one shared mathematical estimator, such as an Iterated Error-State Kalman Filter (IESKF), systems can de-skew LiDAR point clouds in real-time using high-frequency IMU predictions \cite{pan2024lidar}. To correct long-term accumulated drift, systems utilize multi-modal loop closure via Pose Graph Optimization (PGO) to bend the trajectory back onto the correct global path.

While tight coupling focuses on dense High-Definition (HD) maps, alternative paradigms like Vehicle-to-Everything (V2X) and Cooperative Architectures for Transportation Systems (CATS) utilize networked spatial abstractions \cite{v2x_emerging, cats_convoy}. Similarly, Model Predictive Trajectory Tracking uses simplified 2D maps and Hybrid A* planning to reduce 3D SLAM's computational load \cite{ros_mpc}. However, relying purely on abstract mathematical representations creates vulnerabilities during network drops or in unmapped terrain. Thus, local, tightly-coupled 3D SLAM generating continuous HD maps remains strictly required for true vehicle autonomy.

\section{Methodology}

\begin{figure}[htbp]
\centerline{\includegraphics[width=0.45\textwidth]{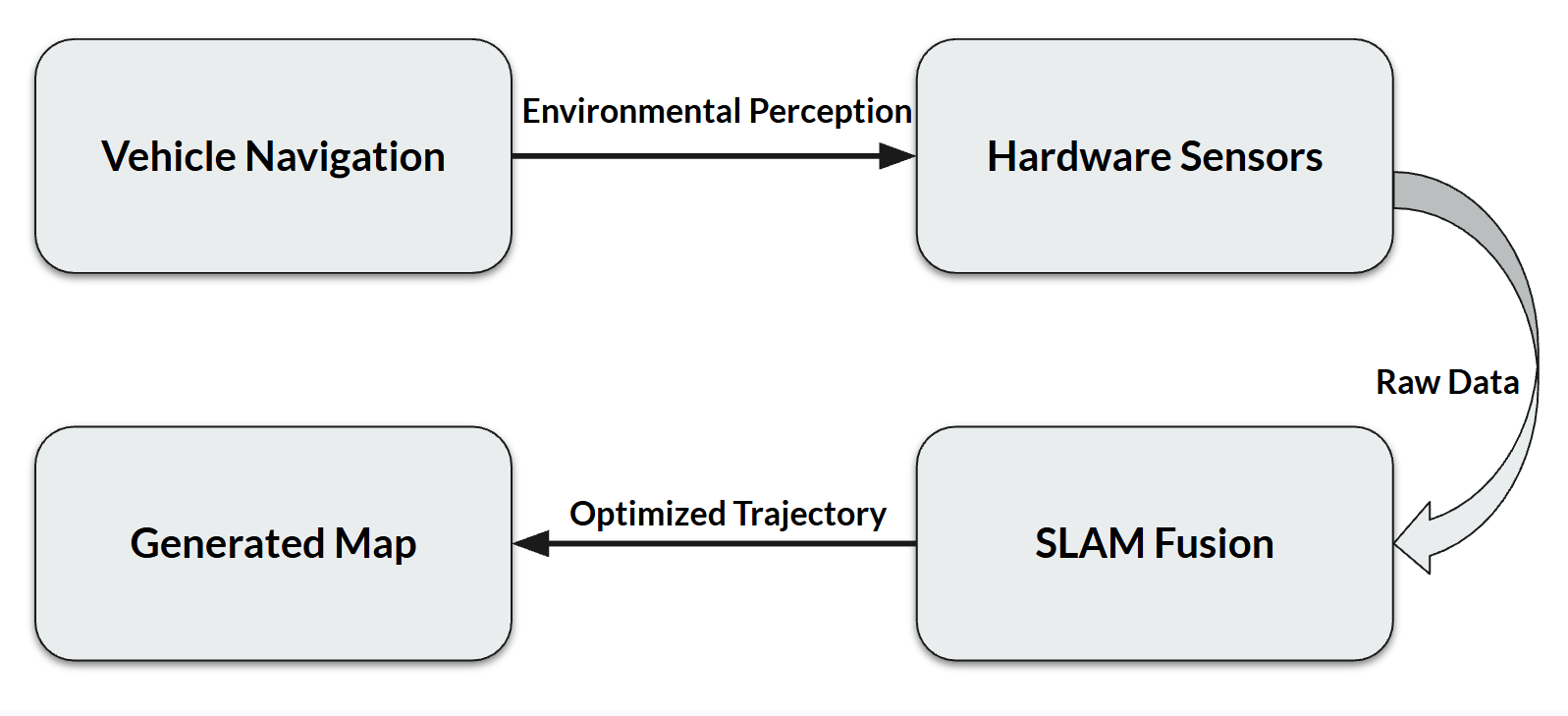}}
\caption{The four-stage SLAM fusion pipeline, demonstrating the flow from raw hardware perception to optimized trajectory and map generation.}
\label{fig:pipeline}
\end{figure}

The mapping system acts as a continuous processing pipeline divided into four distinct stages: (1) Data Collection, (2) Synchronization and Motion Correction, (3) SLAM Processing via IESKF and PGO, and (4) Map Generation utilizing an incremental \textit{ikd-tree} (Fig. \ref{fig:pipeline}).

\subsection{Hardware Platform and Sensors}
The system is built entirely on the Robot Operating System 2 (ROS 2 Humble) middleware. All software runs locally on a Lenovo Legion 5 laptop with an Intel Core i7-13650HX processor and 16 GB of RAM, ensuring real-time performance.

The primary mapping sensor is a Velodyne VLP-16, a 16-beam spinning LiDAR with a range of up to 100 m. Motion is tracked by an LSM6DSOX 6-axis IMU mounted on an Arduino Nano RP2040 Connect, which records linear acceleration and rotational rates. Global position is logged via a 1 Hz GPS receiver.

\begin{figure}[htbp]
\centerline{\includegraphics[width=0.45\textwidth]{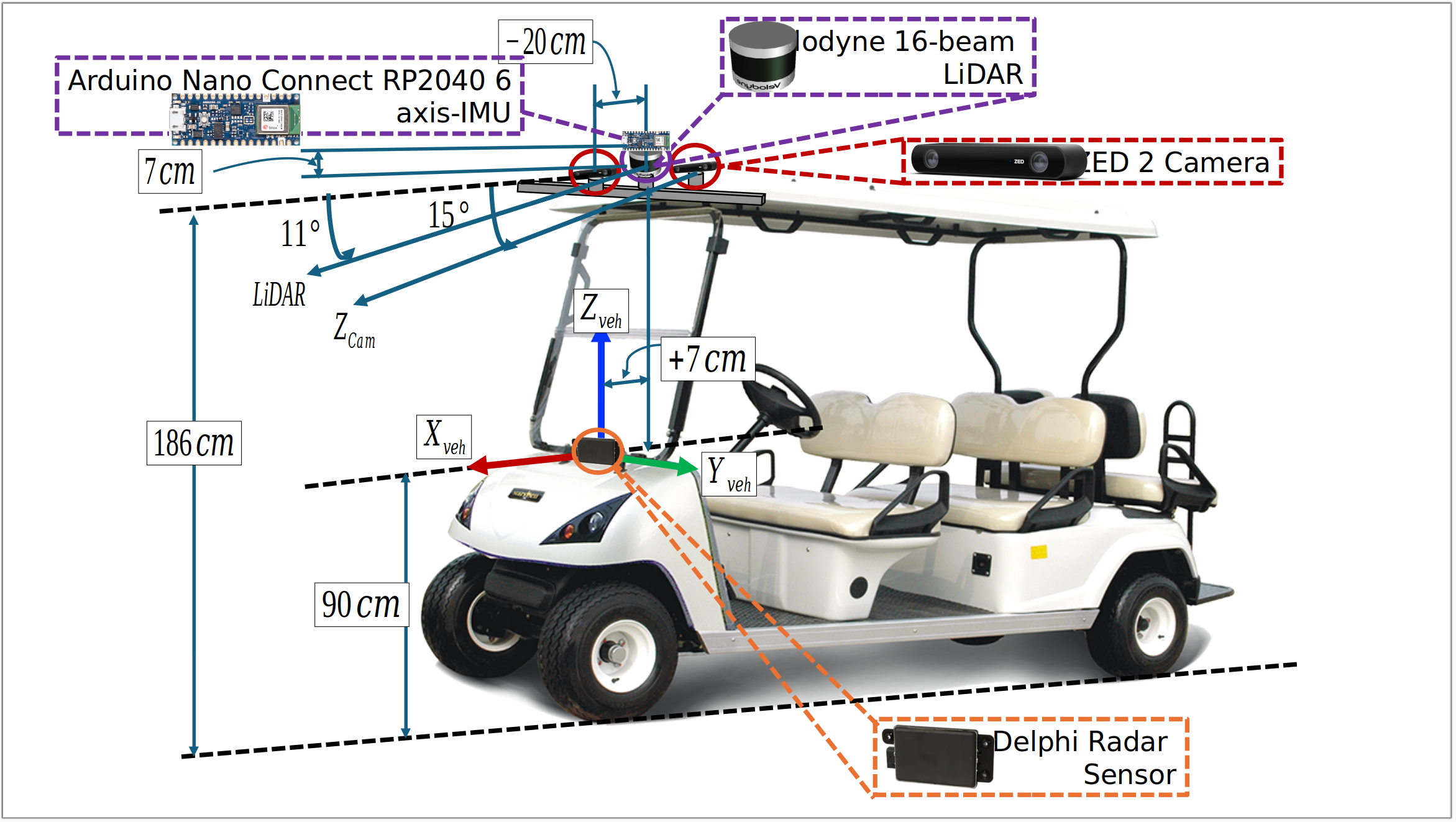}}
\caption{The physical vehicle platform and sensor placement, highlighting the Velodyne LiDAR, IMU, and GPS locations.}
\label{fig:hardware}
\end{figure}

\subsection{ROS 2 Architecture and Data Flow}
To integrate the IMU with ROS 2, a custom serial bridge was developed. Firmware on the RP2040 samples the accelerometer ($\pm 4g$) and gyroscope ($\pm 500$ dps), streaming a 200 Hz serial output. A ROS 2 \texttt{imu\_serial} bridge node reads this stream, recovers SI units ($m/s^2$ and $rad/s$), and publishes standard \texttt{sensor\_msgs/Imu} messages to \texttt{/imu/data}. Velodyne 3D laser scans are published to \texttt{/velodyne\_points} at 10 Hz. 

To ensure spatial consistency, an exact Transform Frame (TF) tree was configured to translate the local vehicle frame to the global map frame. On the real hardware platform, the Velodyne is mounted exactly 1.86 m above the vehicle's footprint (\textit{base\_link}) and pitched $11^\circ$ downward. A static transform applies this height offset and pitch correction to every incoming laser point before it reaches the SLAM algorithm, preventing the algorithm from misinterpreting sloping ground immediately in front of the vehicle as a vertical wall.

\subsection{Test Environments and Algorithm Combinations}
To isolate coordinate frame errors safely before handling complex 3D state estimation, the system was evaluated across three specific environments: \textbf{E-sim} (Gazebo simulation for repeatable exact ground truth), \textbf{E-bag} (offline replay of campus rosbags scored against GPS tracks), and \textbf{E-lab} (live physical hardware checks). 

The system was evaluated in three architectural combinations: \textbf{C1 (TurtleBot3 Baseline)} utilizing a standard 2D differential-drive robot in Gazebo via SLAM Toolbox; \textbf{C2 (2D Velodyne Pipeline)} utilizing the physical vehicle's Velodyne mathematically reduced to a single horizontal ring with heavily tuned parameters to suppress rotational drift; and \textbf{C3 (3D Tightly Coupled Fusion)} fusing full 3D Velodyne point clouds and 200 Hz IMU data via the FAST-LIO2 front-end and GTSAM back-end.

\subsection{FAST-LIO2 and Pose Graph Optimization}
For C3, FAST-LIO2 serves as the odometry front-end, running an Iterated Error-State Kalman Filter (IESKF). In the predict step, it integrates the IMU at 200 Hz to roll the state forward. Each incoming 10 Hz LiDAR scan is de-skewed using this motion, then matched to the map through point-to-plane residuals to correct the predicted state and IMU biases. Key FAST-LIO2 parameters included a \textit{cube\_side\_length} of 1000.0 m and a \textit{det\_range} of 100.0 m.

Because local odometry drifts, a GTSAM back-end acts as a global corrector. The trajectory is modeled as a factor graph optimized via iSAM2. Keyframes are added every 1.5 m. To find a loop, the system searches for an earlier keyframe within 20 m that lies at least 8 keyframes back. Candidates are verified using Iterative Closest Point (ICP), requiring a fitness score exceeding 0.50. 

GPS fixes enter the graph as global constraints. Latitude and longitude are projected onto a flat local plane centered on the first reading using an equirectangular projection. GPS factors use an adaptive noise model scaling from 1.0 m (strong signal) to 5.0 m (weak signal), ensuring a degraded GPS fix cannot forcefully drag an accurate LiDAR map out of alignment.

\section{Experimental Setup and Results}

\subsection{Evaluation Metrics and Limitations}
Every run is scored using the \texttt{evo} toolkit. Absolute Pose Error (APE) measures how far the estimated path lies from the reference using SE(3) Umeyama alignment, capturing global consistency. Relative Pose Error (RPE) measures local accuracy by comparing how the vehicle moved over short fixed segments, capturing local drift.

A critical limitation of the real-world evaluation is that the commercial GPS unit itself drifts by approximately one meter, making it an imperfect reference point. Reported trajectory errors on physical logs are interpreted with this hardware constraint in mind.

\subsection{Phase 1: Simulation Benchmarks (E-sim)}
The initial evaluation phase validated the pipelines within the controlled Gazebo simulation. The C1 TurtleBot3 baseline produced an APE RMSE of 0.011 m and an RPE RMSE of 0.012 m, confirming the foundational mapping architecture was sound.

\begin{figure}[H]
\centering
\begin{minipage}{0.31\columnwidth}
  \centering
  \includegraphics[width=\linewidth]{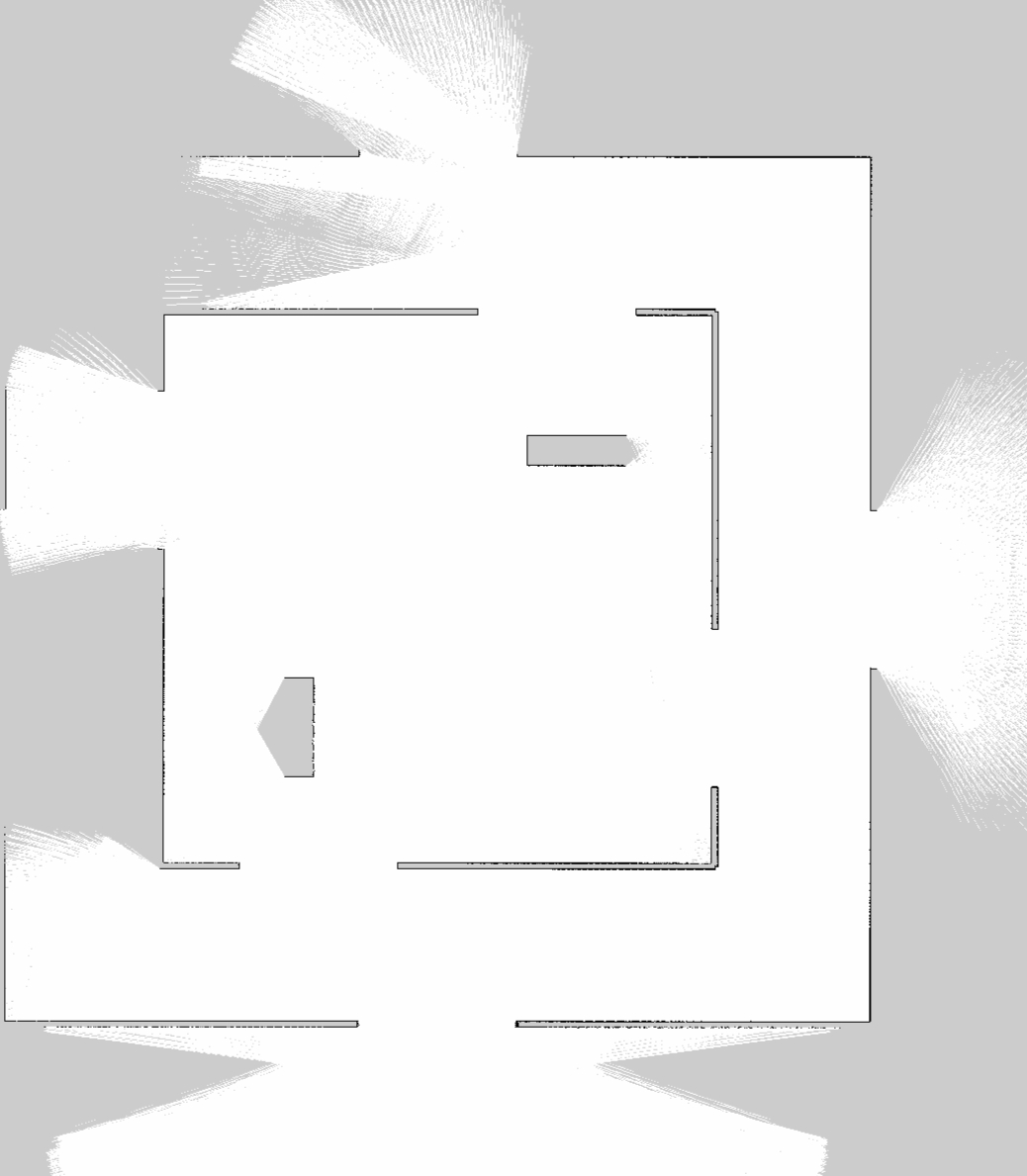}
  \\(a) Velodyne
\end{minipage}\hfill
\begin{minipage}{0.31\columnwidth}
  \centering
  \includegraphics[width=\linewidth]{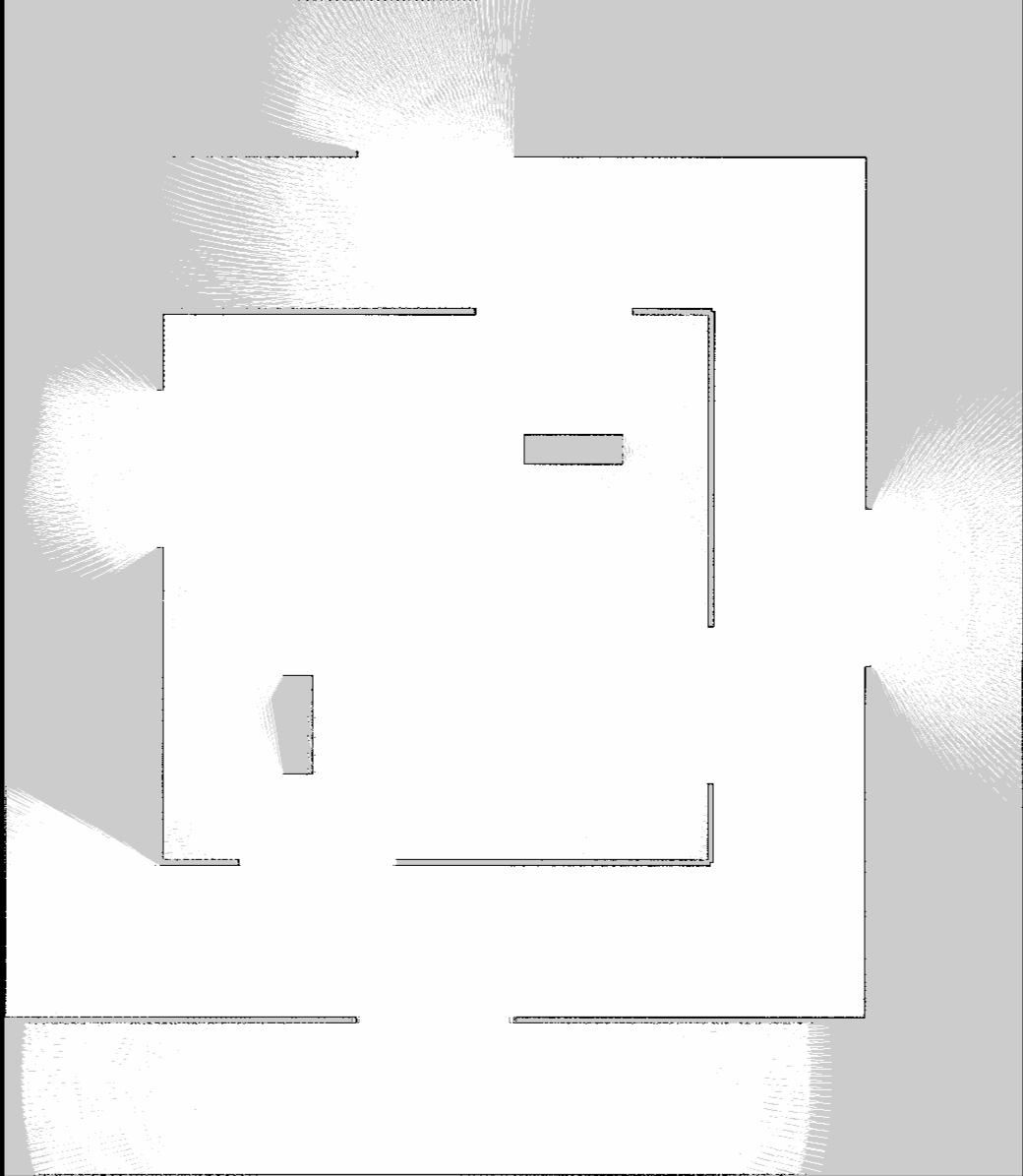}
  \\(b) RoboSense
\end{minipage}\hfill
\begin{minipage}{0.27\columnwidth}
  \centering
  \includegraphics[width=\linewidth]{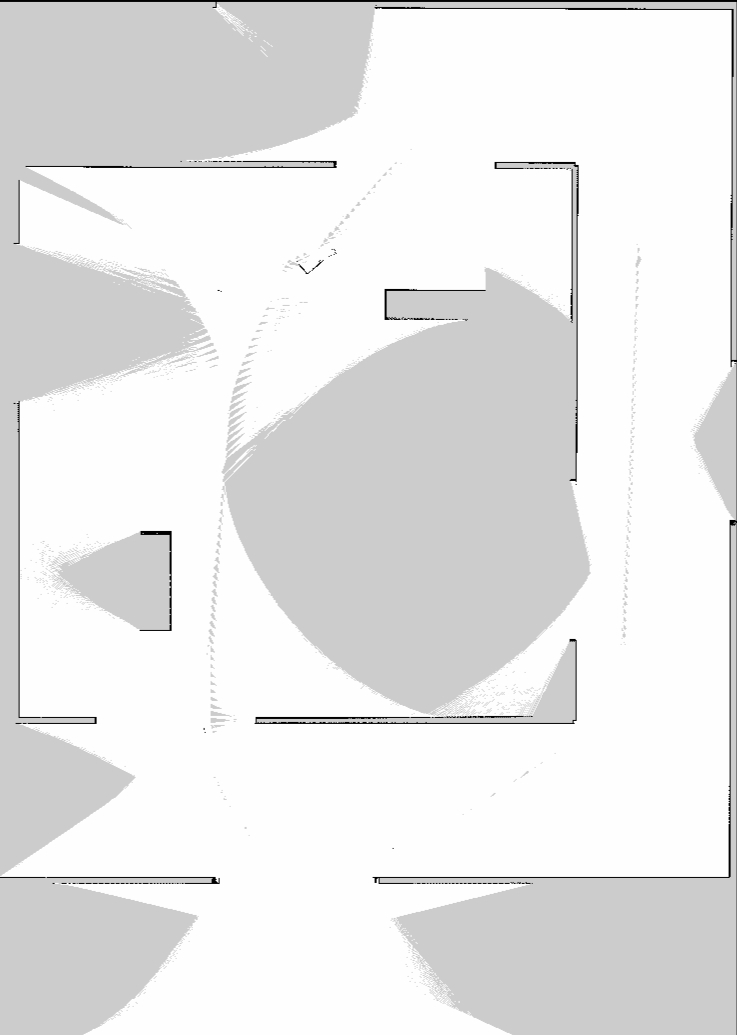}
  \\(c) RPLidar
\end{minipage}
\caption{2D occupancy grids using 2D SLAM Toolbox baseline across three different sensor profiles in E-sim.}
\label{fig:occupancy}
\end{figure}

For the C2 pipeline, three LiDAR scanners were compared in simulation using the exact same SLAM Toolbox configuration. As shown in Table \ref{tab:sensor_comparison} and visualized in Fig. \ref{fig:occupancy}, the single-ring Velodyne achieved the highest accuracy.

\begin{table}[htbp]
\caption{Comparison of LiDAR Sensors via SLAM Toolbox (E-sim)}
\centering
\begin{tabularx}{\columnwidth}{X c c}
\toprule
\textbf{Sensor Type} & \textbf{APE RMSE (m)} & \textbf{RPE RMSE (m)} \\
\midrule
Velodyne VLP-16 (Single) & 0.027 & 0.016 \\
RoboSense Airy & 0.089 & 0.028 \\
RPLidar A2 & 0.203 & 0.067 \\
\bottomrule
\end{tabularx}
\label{tab:sensor_comparison}
\end{table}
\vspace{-15pt}

\begin{figure}[h]
\centerline{\includegraphics[width=0.17\textwidth]{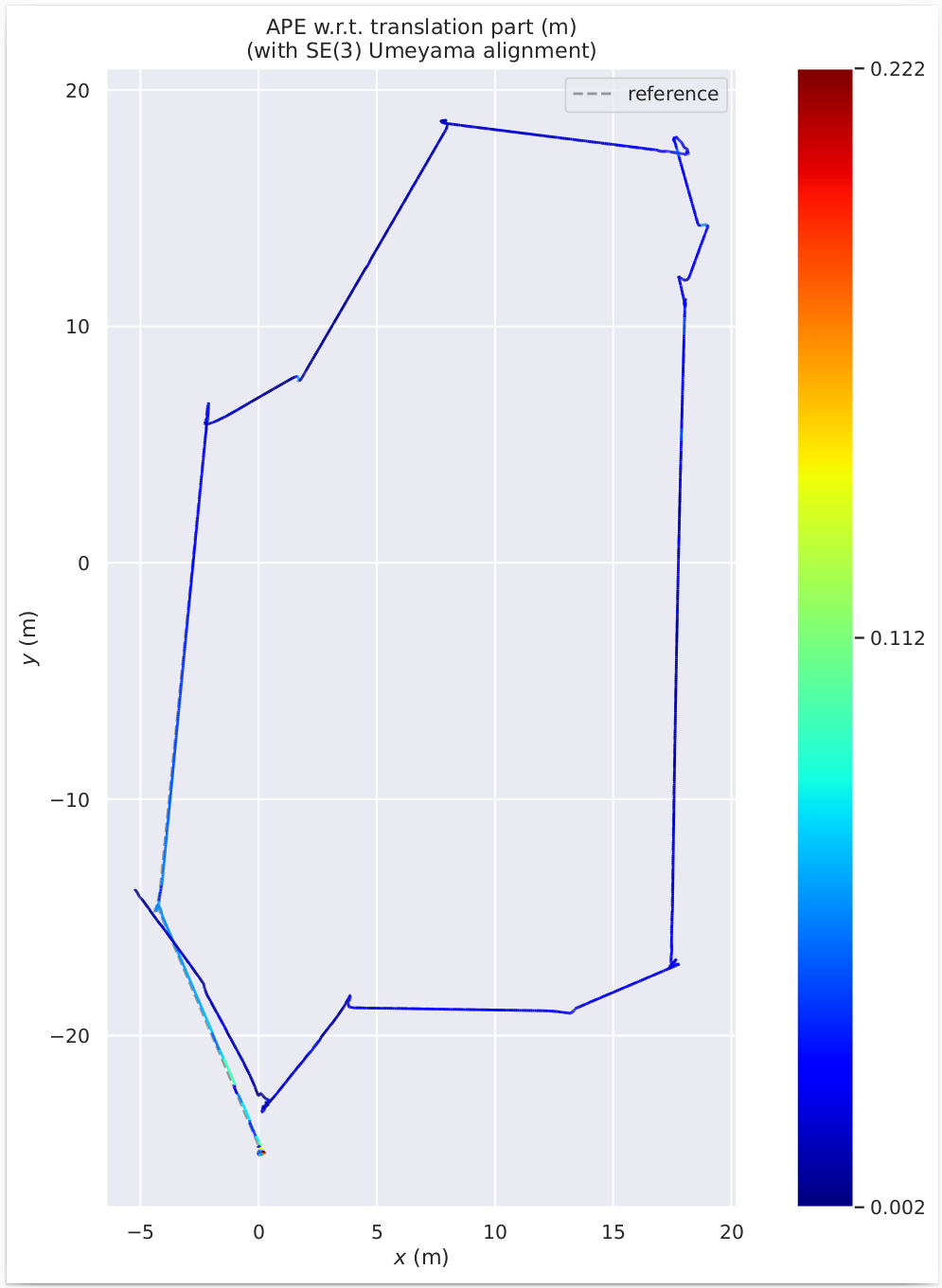}}
\caption{Estimated C2 trajectory evaluated against the ground truth in the Gazebo simulation, colored by absolute error magnitude along the path.}
\label{fig:evo_sim_plot}
\end{figure}

The low global error of the Velodyne configuration is  plotted against the simulation ground truth in Fig. \ref{fig:evo_sim_plot}.

Upgrading to the full C3 tightly-coupled architecture, the open-loop FAST-LIO2 front-end achieved highly accurate 3D local odometry in simulation (Table \ref{tab:c3_results}).

\begin{table}[htbp]
\caption{Pose Error for C3 FAST-LIO2 Front-End (E-sim)}
\centering
\begin{tabularx}{\columnwidth}{X c c c c}
\toprule
\textbf{Metric} & \textbf{Mean} & \textbf{Median} & \textbf{RMSE} & \textbf{Max} \\
\midrule
APE (m) & 0.580 & 0.535 & 0.615 & 1.308 \\
RPE (m) & 0.108 & 0.067 & 0.154 & 0.551 \\
\bottomrule
\end{tabularx}
\label{tab:c3_results}
\end{table}

\subsubsection{GTSAM Loop Closure Behavior}
\begin{figure}[h]
\centering
\begin{minipage}{0.43\columnwidth}
  \centering
  \includegraphics[width=\linewidth]{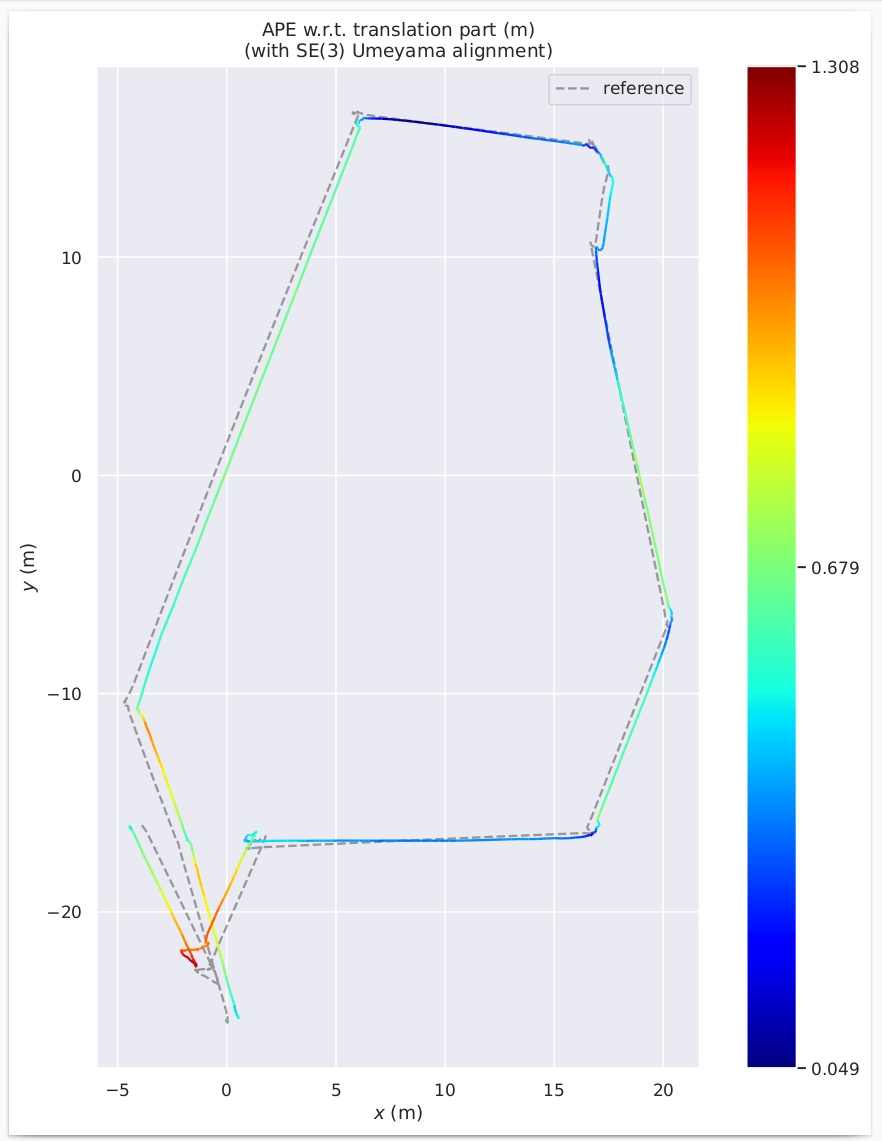}
  \\(a) Before PGO
\end{minipage}\hfill
\begin{minipage}{0.43\columnwidth}
  \centering
  \includegraphics[width=\linewidth]{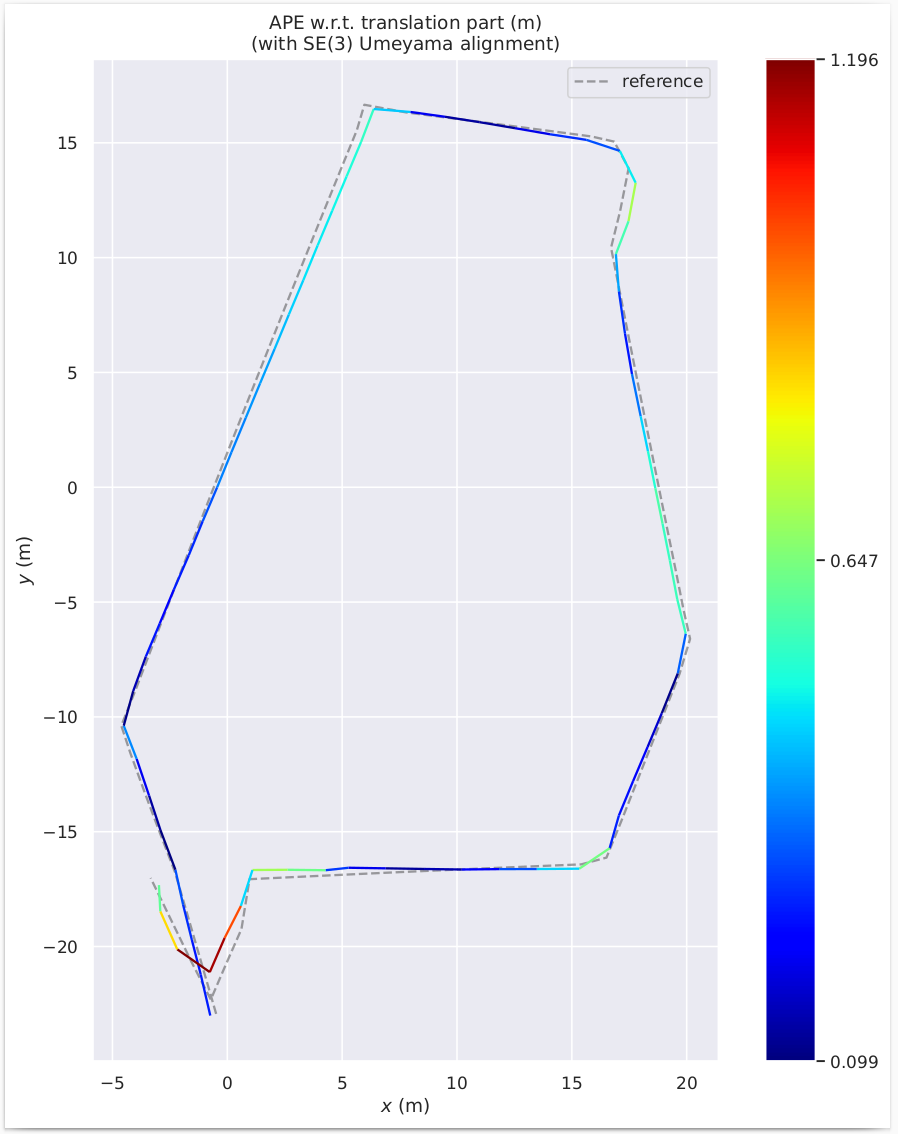}
  \\(b) After PGO
\end{minipage}
\caption{Effect of the GTSAM pose-graph loop closure in simulation: (a) open-loop trajectory versus (b) the globally re-optimized trajectory.}
\label{fig:loop_closure}
\end{figure}

To isolate the PGO back-end, the C3 simulation was processed with loop closure enabled and disabled (Table \ref{tab:loop_closure} and Fig. \ref{fig:loop_closure}). When applied in clean simulation, the APE dropped to 0.460 m, but the RPE worsened to 0.276 m. Because the true drift in pristine simulation is much lower than the fixed noise model (sigmas of $\sim0.1$ m), the optimizer treats accurate odometry as noisier than loop constraints, bending smooth local motion. 

\begin{table}[H]
\caption{Effect of GTSAM Loop Closure on Pose Error (E-sim)}
\centering
\begin{tabularx}{\columnwidth}{l l c c}
\toprule
\textbf{Sensor} & \textbf{Configuration} & \textbf{APE RMSE} & \textbf{RPE RMSE} \\
\midrule
Velodyne & Open Loop & 0.615 m & 0.154 m \\
Velodyne & + GTSAM PGO & 0.460 m & 0.276 m \\
\midrule
RoboSense & Open Loop & 0.404 m & 0.152 m \\
RoboSense & + GTSAM PGO & 0.585 m & 0.297 m \\
\bottomrule
\end{tabularx}
\label{tab:loop_closure}
\end{table}

\begin{figure}[H]
\centering
\begin{minipage}{0.45\columnwidth}
  \centering
  \includegraphics[width=\linewidth]{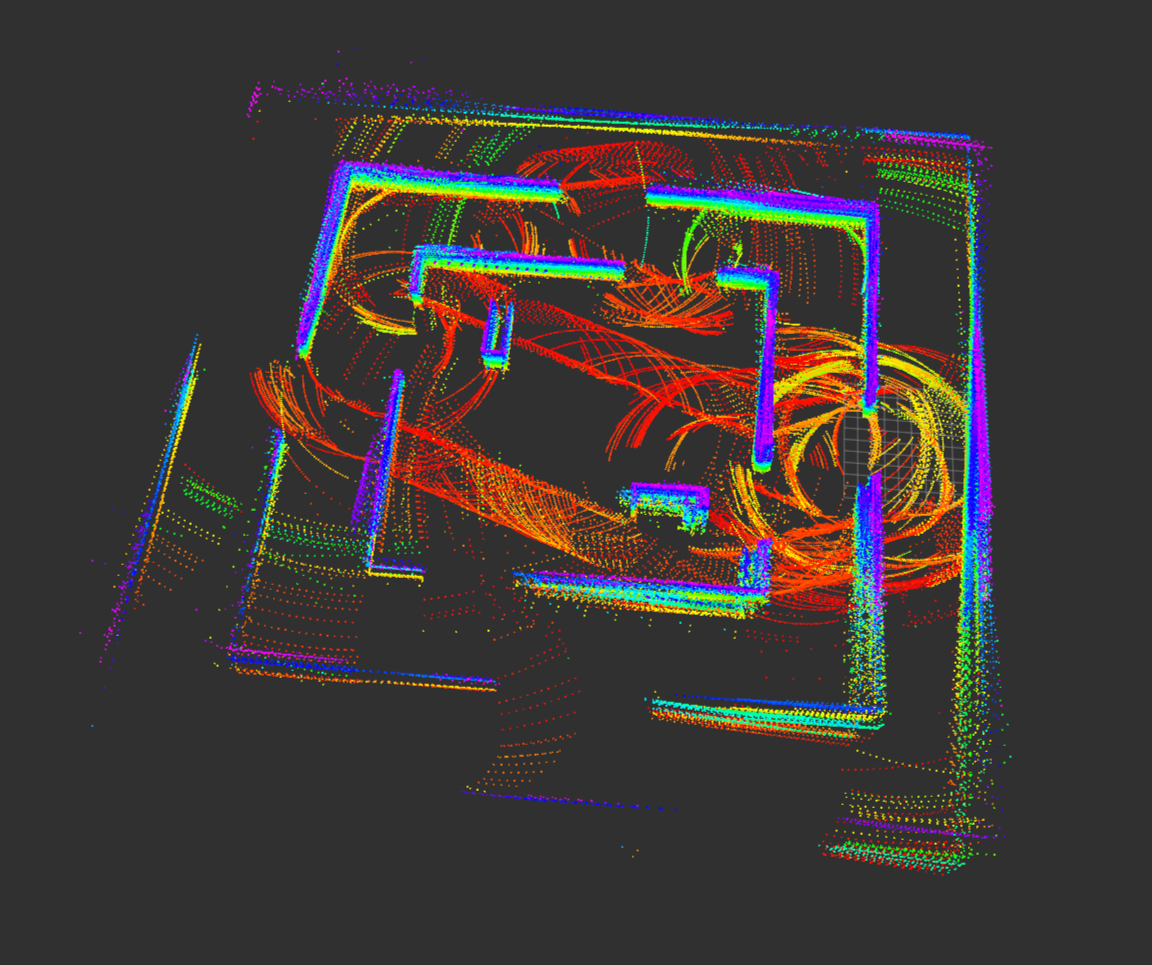}
  \\(a) Velodyne Map
\end{minipage}\hfill
\begin{minipage}{0.47\columnwidth}
  \centering
  \includegraphics[width=\linewidth]{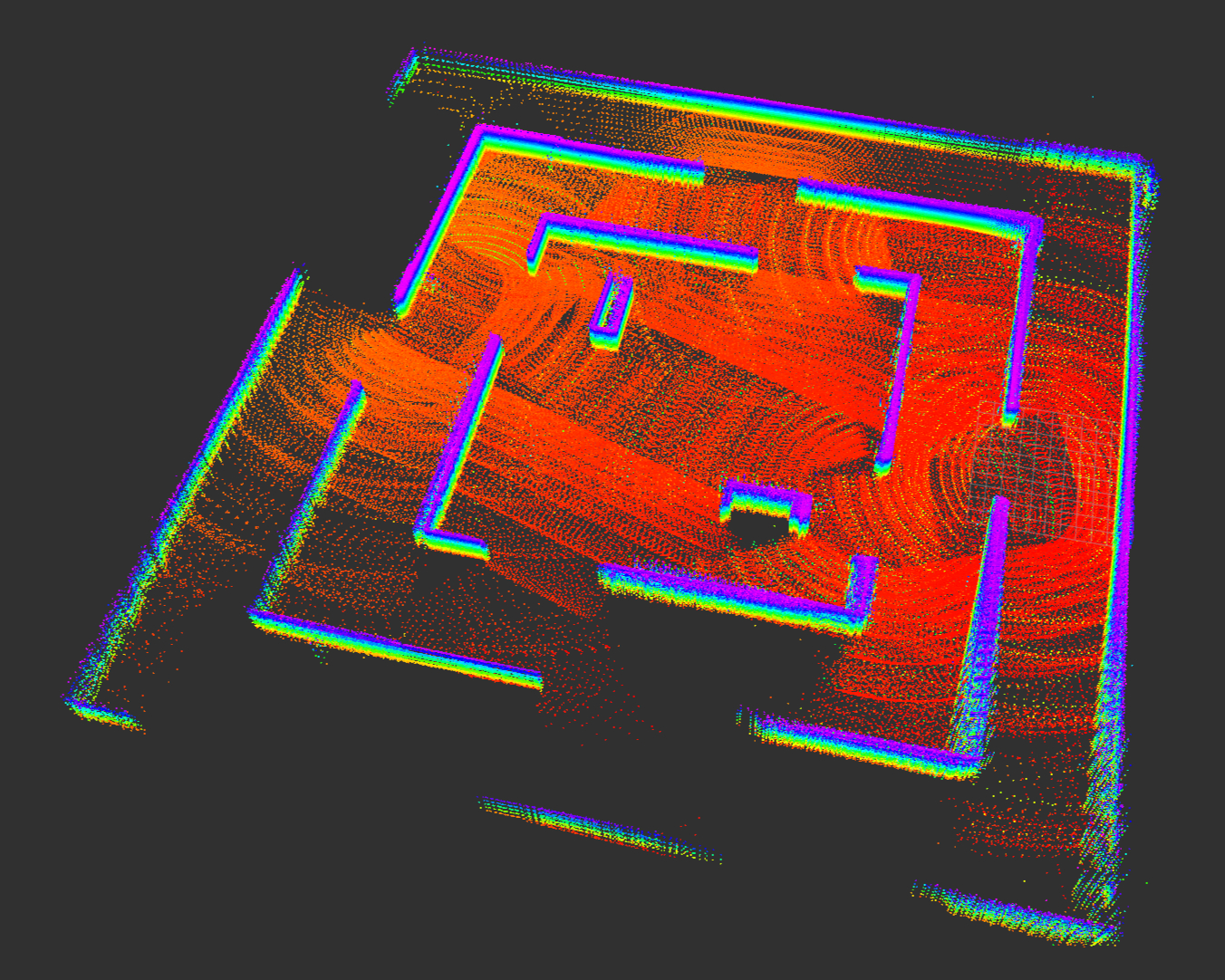}
  \\(b) RoboSense Map
\end{minipage}
\caption{Three-dimensional point-cloud maps generated in simulation using FAST-LIO2, comparing the structural density of the different scanners.}
\label{fig:sim_3d_maps}
\end{figure}

Table \ref{tab:cross_comp} brings the combinations together for direct comparison. The 2D combinations (C1, C2) report the lowest absolute error, expected as they solve a constrained planar problem with reliable odometry. The 3D system (C3) carries a larger absolute error because it estimates a full 6-DOF trajectory from LiDAR and inertial data alone, returning a far richer map in exchange.
\vspace{-10pt}
\begin{table}[H]
\caption{Cross-Combination Comparison on Matched Metrics (E-sim)}
\centering
\begin{tabularx}{\columnwidth}{X c c c}
\toprule
\textbf{ID} & \textbf{Map Type} & \textbf{APE RMSE (m)} & \textbf{RPE RMSE (m)} \\
\midrule
C1 & 2D Occupancy & 0.011 & 0.012 \\
C2 & 2D Occupancy & 0.027 & 0.016 \\
C3 & 3D Point Cloud & 0.615 & 0.154 \\
\bottomrule
\end{tabularx}
\label{tab:cross_comp}
\end{table}

\subsection{Phase 2: Real-World Deployments (E-lab \& E-bag)}
\subsubsection{Live 2D Hardware Tests and Rosbag Replay (C2)}
\begin{figure}[H]
\centering
\begin{minipage}{0.47\columnwidth}
  \centering
  \includegraphics[width=\linewidth]{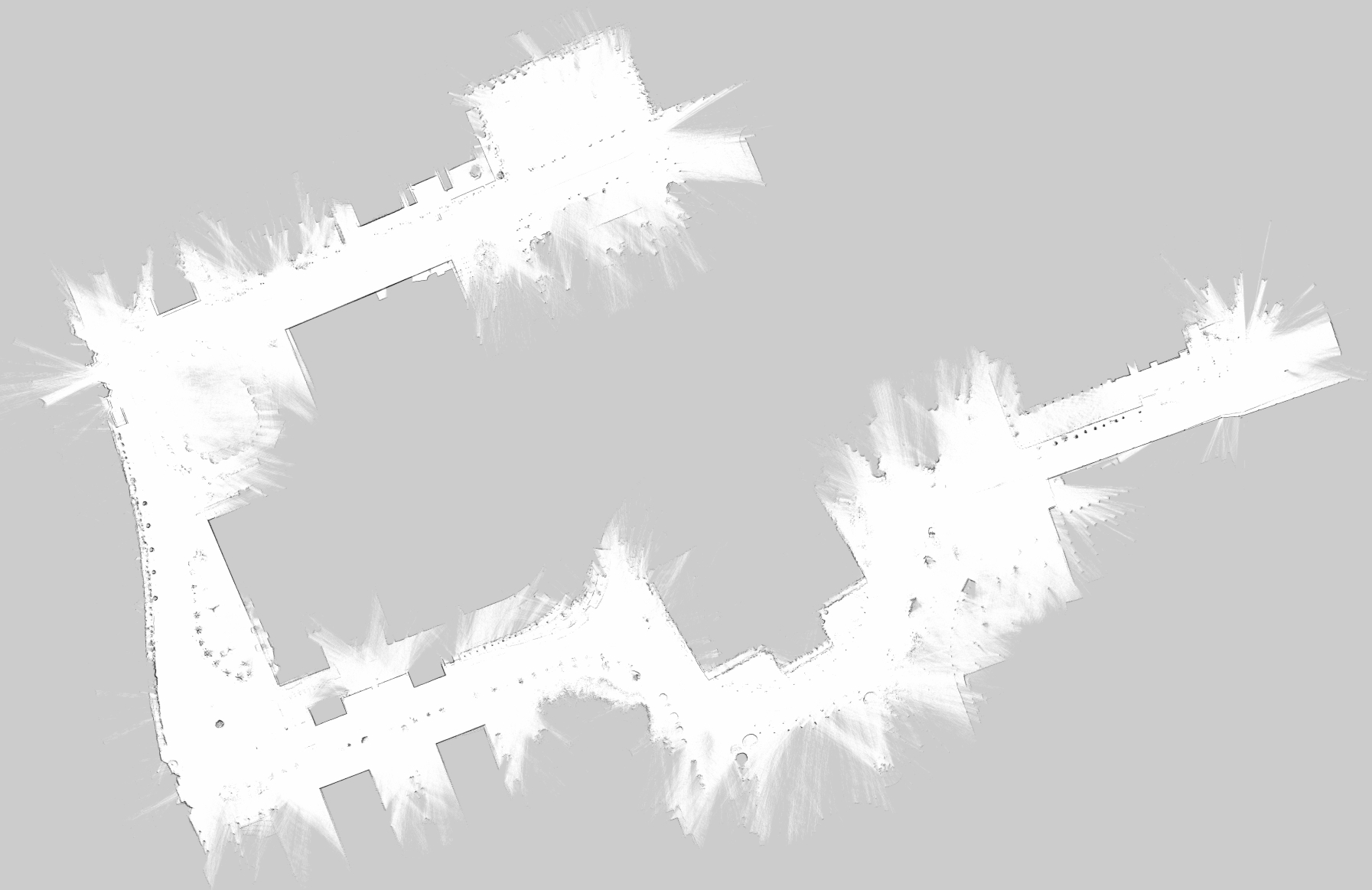}
  \\(a) Cycle 1 Map
\end{minipage}\hfill
\begin{minipage}{0.45\columnwidth}
  \centering
  \includegraphics[width=\linewidth]{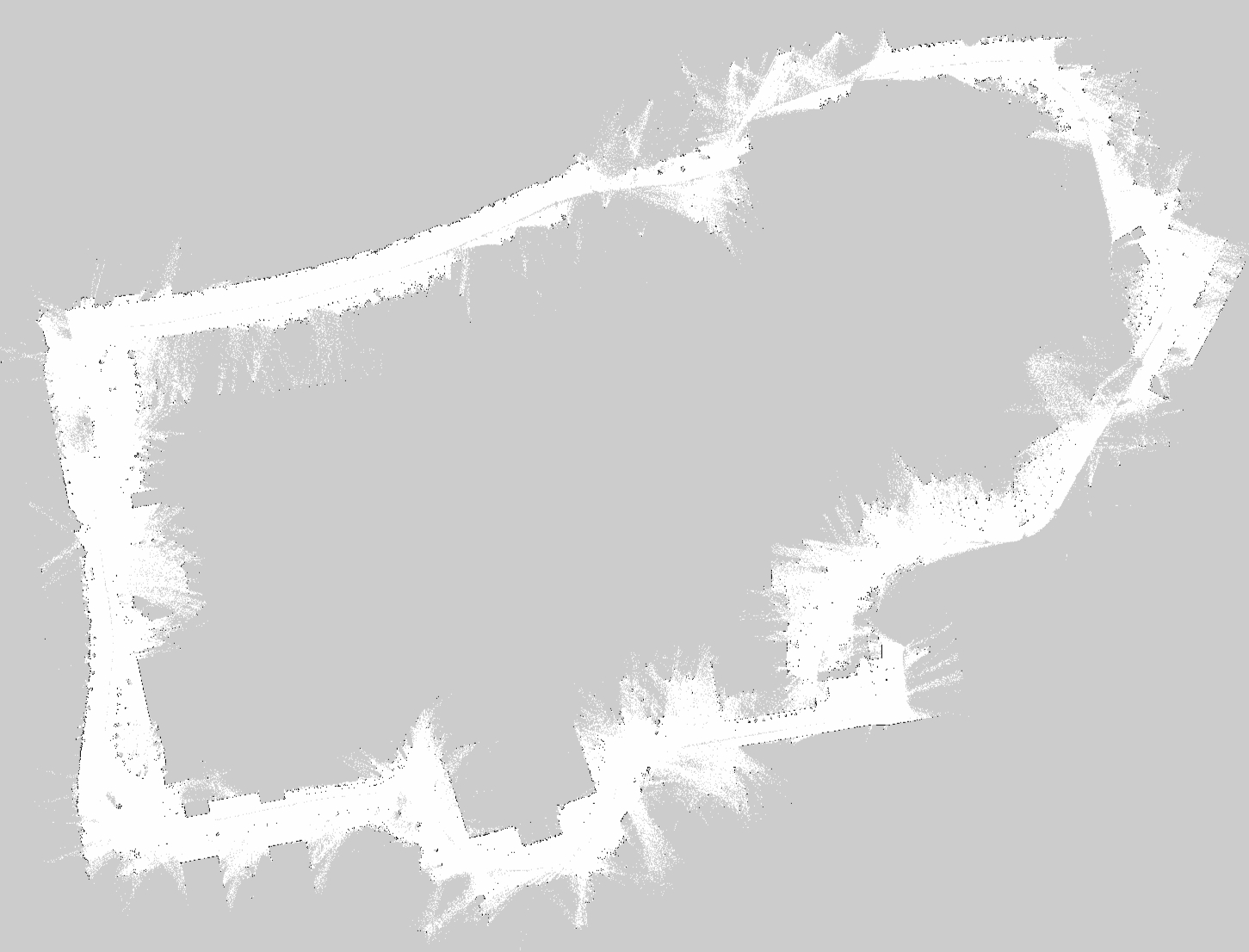}
  \\(b) Cycle 2 Map
\end{minipage}
\caption{Two-dimensional real-world occupancy grids generated by the C2 pipeline for (a) Cycle 1 and (b) Cycle 2.}
\label{fig:real_world_2d}
\end{figure}

\begin{figure}[h]
\centering
\begin{minipage}{0.45\columnwidth}
  \centering
  \includegraphics[width=\linewidth]{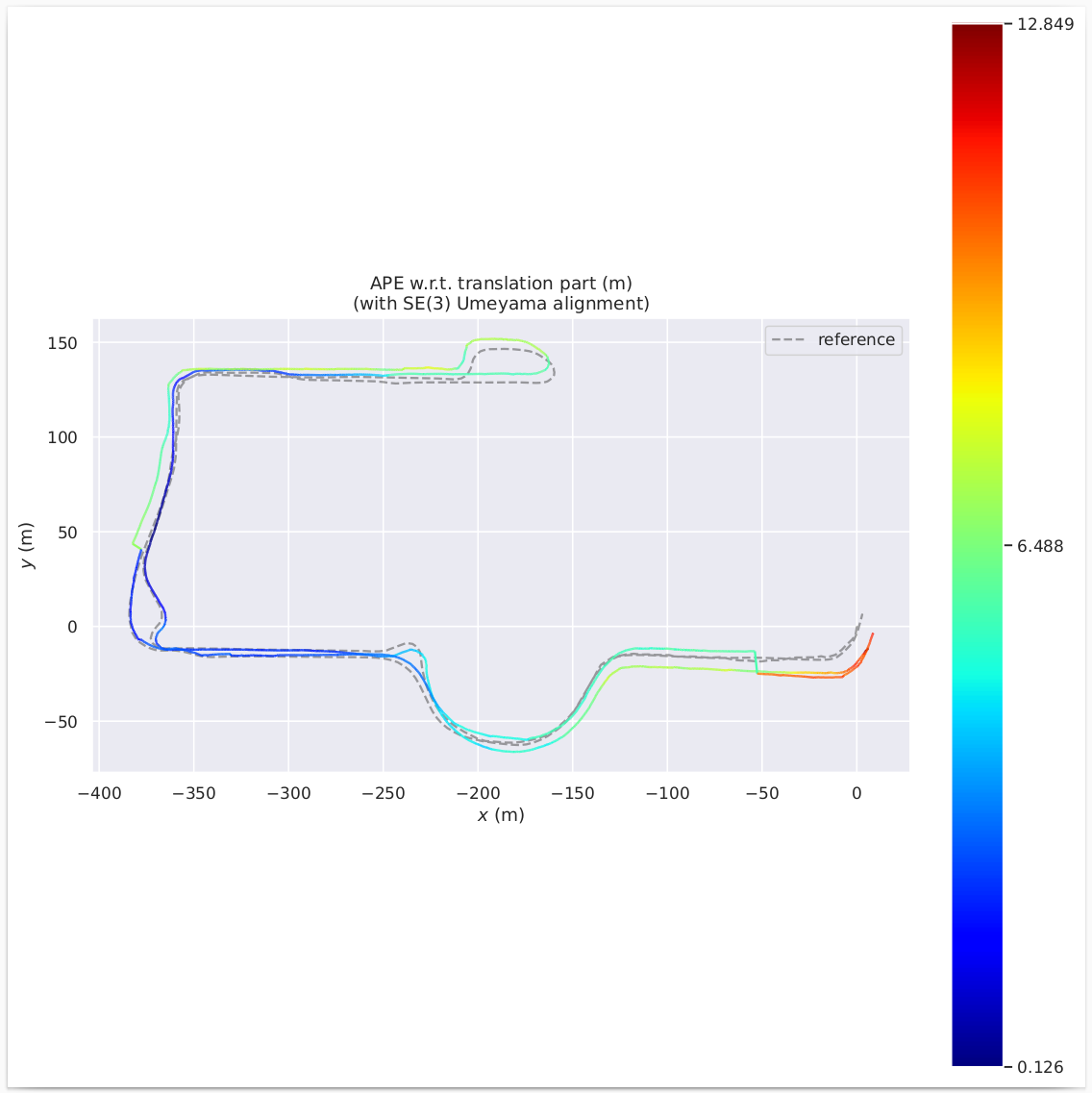}
  \\(a) Cycle 1 Error
\end{minipage}\hfill
\begin{minipage}{0.49\columnwidth}
  \centering
  \includegraphics[width=\linewidth]{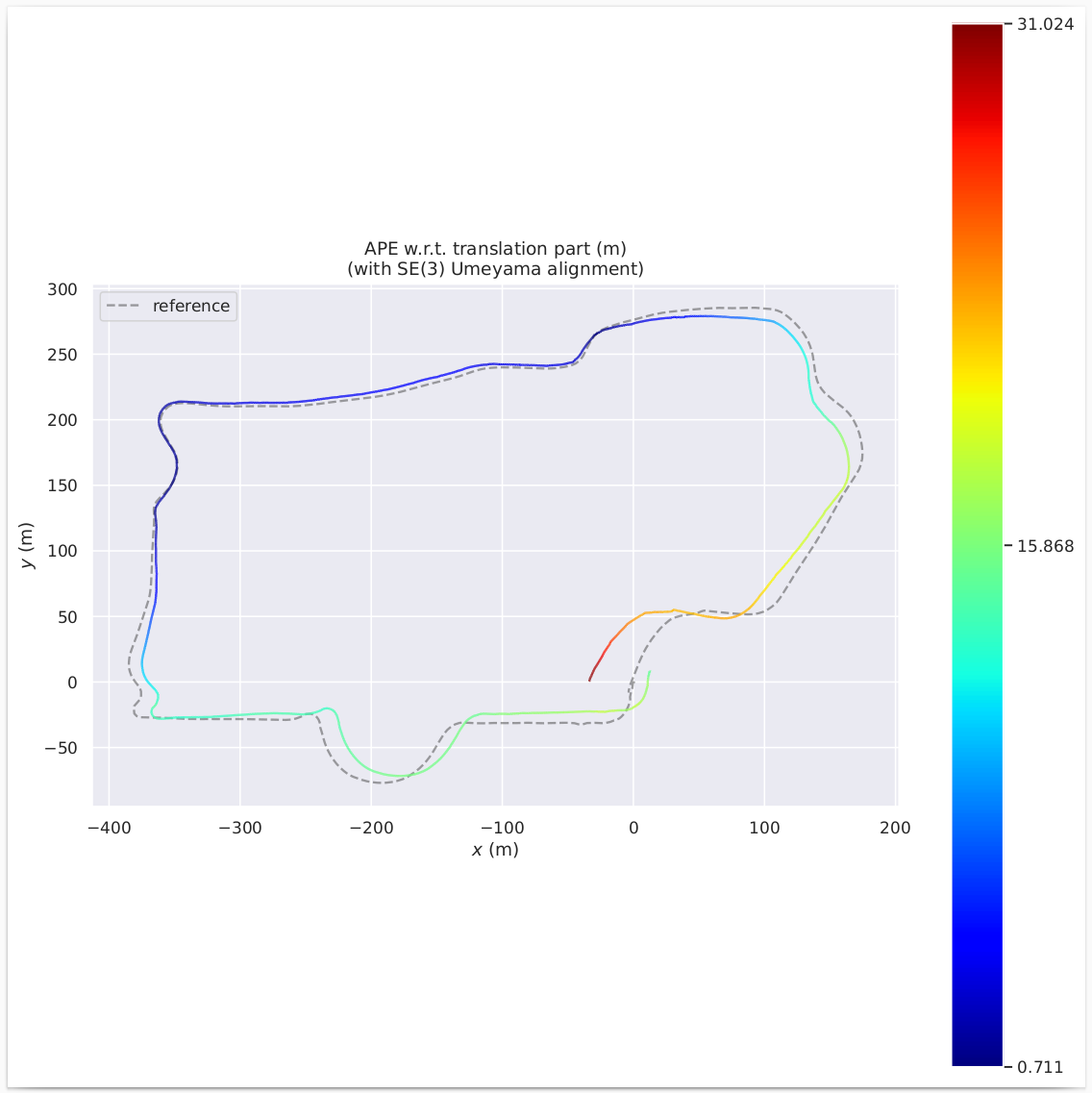}
  \\(b) Cycle 2 Error
\end{minipage}
\caption{Absolute Pose Error (APE) trajectories for the C2 pipeline evaluated against GPS ground truth during real-world campus navigation.}
\label{fig:real_world_traj}
\end{figure}

The Velodyne was first run live (E-lab) to confirm the SLAM Toolbox pipeline correctly generated local occupancy grids. The pipeline was then evaluated against two outdoor campus rosbags (E-bag): Cycle 1 (522 position fixes) and Cycle 2 (a 560 m by 363 m loop). Transitioning to unconstrained environments highlighted 2D scanning limitations. While C2 achieved high local consistency (RPE RMSE 0.206 m and 0.096 m), the resulting maps (Fig. \ref{fig:real_world_2d}) and trajectories (Fig. \ref{fig:real_world_traj}) suffered massive global drift (APE RMSE 5.805 m and 13.677 m). The sparse single ring fails to constrain heading effectively outdoors, causing radial warp and necessitating a 3D approach.

\subsubsection{Live 3D Hardware Tests (C3)}
\begin{figure}[h]
\centerline{\includegraphics[width=0.3\textwidth]{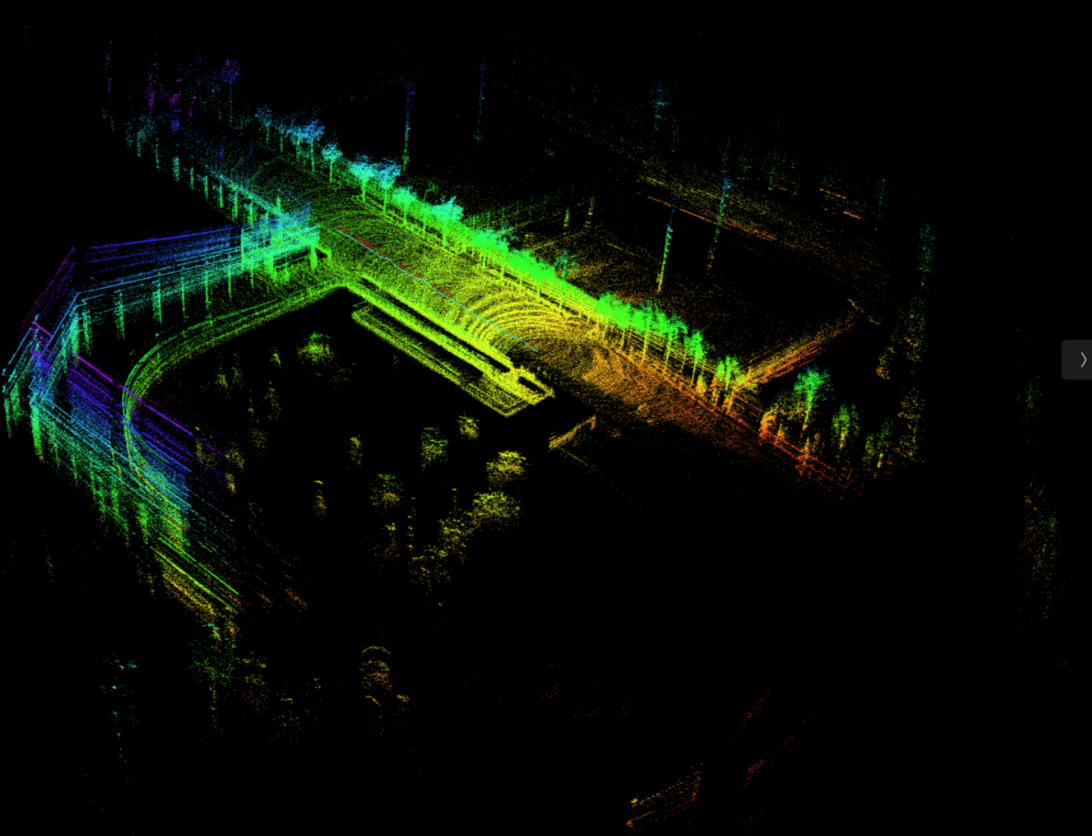}}
\caption{3Dl point-cloud map produced by the FAST-LIO2 pipeline during live physical navigation on campus.}
\label{fig:point_cloud}
\end{figure}

To overcome the 2D limitations, the full FAST-LIO2 3D system was deployed live on the physical vehicle (E-lab). During live campus navigation, the tightly-coupled system functioned smoothly, successfully generating dense 3D point-cloud maps (Fig. \ref{fig:point_cloud}) and robust state estimations in real-time without computational bottlenecking.

\subsubsection{Deferred Real-Hardware Replay}
While live hardware tests confirmed real-time operation, offline quantitative evaluation of C3 via \texttt{evo} on the Cycle 1 and Cycle 2 logs was deferred. The recorded campus rosbags logged the LiDAR streams but critically lacked the \texttt{/imu/data} stream. Because FAST-LIO2 strictly requires a high-rate inertial stream for deskewing, C3 cannot be reconstructed from these specific recordings. 

\subsection{Discussion of Failure Modes}
Testing revealed several characteristic failure modes. \textbf{1. Sparseness of the Single Ring:} In the C2 combination, long stretches of visually similar corridors gave the 2D scan matcher little to lock onto, integrating spurious rotations into severe radial drift and justifying the C3 architecture. \textbf{2. False-Positive Loop Closures:} In highly repetitive regions, two distinct locations can produce point clouds that align with high ICP fitness scores. Tightening keyframe-selection parameters and enforcing a strict pose-consistency check successfully suppressed false positives across our tests.

\section{Conclusion and Future Work}
This research implemented a tightly-coupled multi-sensor SLAM architecture. Utilizing a phased methodology proved critical: employing a 2D SLAM Toolbox baseline isolated coordinate transform errors before deploying the mathematically intensive 3D FAST-LIO2 pipeline. The results demonstrate that high-frequency IMU deskewing combined with 3D LiDAR data generates resilient HD point-clouds, overcoming vulnerabilities inherent to relying strictly on abstract mathematical map abstractions. Future work will focus on complete offline replays of campus rosbags featuring perfectly synchronized, high-rate IMU data to allow dynamic tuning of GTSAM factor sigmas.

\end{document}